\documentclass[11pt,a4paper]{article}
\usepackage[utf8]{inputenc}
\usepackage[hyperref]{acl}
\usepackage{times}
\usepackage{latexsym}

\usepackage{microtype}

\usepackage{paralist}
\usepackage{amsmath}
\usepackage[subtle]{savetrees}
\usepackage{graphicx}
\usepackage{csquotes}
\usepackage{float}

\usepackage{fancyhdr} 
\title{Zeitenwenden: Detecting changes in the German political discourse}

\author{Kai-Robin Lange \and Jonas Rieger \and Niklas Benner \and Carsten Jentsch\\
Department of Statistics, TU Dortmund University, 44221 Dortmund, Germany\\
\texttt{\{kalange, rieger, benner, jentsch\} @statistik.tu-dortmund.de}\\}

\begin{document}
\maketitle
\thispagestyle{fancy} 
\begin{abstract}
From a monarchy to a democracy, to a dictatorship and back to a democracy -- the German political landscape has been constantly changing ever since the first German national state was formed in 1871. After World War II, the Federal Republic of Germany was formed in 1949. Since then every plenary session of the German Bundestag was logged and even has been digitized over the course of the last few years. We analyze these texts using a time series variant of the topic model LDA to investigate which events had a lasting effect on the political discourse and how the political topics changed over time. This allows us to detect changes in word frequency (and thus key discussion points) in political discourse.
\end{abstract}

\section{Introduction}
\enquote{Wir erleben eine Zeitenwende} -- \enquote{We are witnessing a turn of eras}. This quote by Germany's chancellor Scholz \cite{Scholz} was the result of a turning point of political discourse in Germany after the outbreak of the Russian-Ukrainian war of 2022. This was an obvious turning point in German politics, as not only the discourse within the parliament, but also the decision making of the German government changed. For instance, 100 billion euros are planned to be spend as additional military expenses \cite{Scholz}. Throughout German history, there have been many changes and turning points in political discourse, but not all of them are as clearly reported or as obvious as this one. Many of them may not have been as clearly remembered because they were gradual rather than rapid changes, or because they did not have an immediate impact on real-world politics.

In this paper, we investigate all 72 years of plenary sessions of the German Bundestag to grasp the development of political discourse in Germany. We analyze these sessions by interpreting them as a time series of textual data for which we use a change detection method proposed by \citet{rieger.t2s}. For this, we use a rolling version of the topic model latent Dirichlet allocation (LDA), which is designed to construct topics that are coherent over time and allows for changing vocabulary. Within the resulting topics, we detect changes by analyzing the actual word usage in topics compared to theoretically expected word usage if no change occurs, which is determined using resampling. With this method, we are also able to differentiate between short-term changes and persistent ones. While "Zeitenwende" is a broad term, we refer to it as a persistent change in the way of how a political topic is discussed. This change in discussion can stem from differing speech patterns or from changing contents of said topic. The former is rarely detected as a change, as a changing speech pattern usually develops slowly. If it does change suddenly, this is, with minor exceptions, due to a change in formality, which is of no interest for our analysis as it does not affect the content of the discussions. The latter however is interesting for our analysis, as it symbolizes that the topic has changed for good due to economic, cultural or diplomatic events or developments. While this change may not affect the entirety of the political spectrum, we consider it to be a Zeitenwende, if it persistently changes one of the 30 most common and important topics of German political discussions, which we analyze using our topic model.

Data sets as large as the German parliament discussions are too large to be read and interpreted all manually. While qualitative expert analysis is needed to analyze German politics over the last decades, a quantitative text analysis can help in this regard by providing experts with ideas for what to look at and by verifying their results with an empirical basis. Our analysis aims to do just that, as the model used allows for changes of different magnitudes to be detected by simply adjusting a single parameter. Experts can use these findings to back up their qualitative results by pointing out the importance and long lasting effect of a change for the political discussions at the time or gather ideas for changes in uncommon topics. By tuning the so-called "mixture"{}-parameter of this model, even rather niche changes can be found to interpret and to compare them based on the tuning needed to detect them (a high tuning parameter indicates a major change, a smaller one indicates a niche change).

\citet{walter2021} use a similar data set to analyze ideological shifts throughout German history. This data set, DeuParl \cite{DeuPARL}, does contain data from plenary sessions since 1867 to 2020. The data before 1949 are however less structured and contain clusters of incoherent text due to being automatically created by scanning old documents, which is why we do not use them in this analysis. While cleaning the data set is a task on its own, political analyses such as this one could greatly benefit from a \enquote{clean} version of this Reichstag data set as it enables to analyze German politics for an even larger period of time.

In a complementary approach, \citet{jentsch2020, jentsch2021} propose a (time-varying) Poisson reduced rank model for party manifestos to extract information on the evolution of party positions and of political debates over time. 

\section{Change detection via rolling modeling}
\label{Methods}
We make use of a rolling version of the classical LDA \cite{blei2003} estimated via Gibbs sampling \cite{griffiths2004}. This method is referred to as RollingLDA \cite{rieger.rolling} and allows new data to be added without manipulating the LDA assignments of the previous model. For this, a more reliable version of the classical LDA is used up to a date \texttt{init}. Then, according to a user-specified periodicity, minibatches of documents (\texttt{chunks}) are modeled using the data available up to that point. Moreover, the model's knowledge of previous documents is constrained in that it only uses the LDA assignments from a given time period (\texttt{memory}) to initialize the modeling of the new texts. Newly occurring vocabulary is added to the model vocabulary and subsequently considered as soon as it occurs more than five times in a minibatch. This flexibility enables the model to adapt for mutations of topics in the form of gradual or abrupt changes in word frequencies.

The minibatches are numbered in ascending order starting with the initialization batch: $t=0, \ldots, T$. Then, using the change detection algorithm by \citet{rieger.t2s}, we get our set of detected changes over time by
\begin{align*}
C_k = \left\{t \mid \cos\left(n_{k|t}, n_{k|(t-z_k^t):(t-1)}\right) < q_{k}^t \right\},
\end{align*}
where $0 < t \leq T$ refers to a specific minibatch and $k \in \{1, \ldots K\}$ to one topic. As proposed by \citet{rieger.t2s}, $q_k^t \in [0,1]$ denotes the $0.01$ quantile of the set of cosine similarities when $n_{k|t}$ is replaced by $\tilde{n}_{k|t}^r, r = 1, \ldots, 500$, where $\tilde{n}_{k|t}^r$ denotes a resampled frequency vector under expected change and $n_{k|t}$ the observed vocabulary frequencies for each topic; analogously $n_{k|(t-z_k^t):(t-1)}$ refers to the sum of the count vectors from time points $t-z_k^t$ until $t-1$. The algorithm has two parameters: the maximum length of the reference period to compare to, $z_k^{\text{max}}$, and the intensity of the expected change under normal conditions $p$. Using the mixture-parameter $p \in [0,1]$, which can be tuned based on how substantial the detected changes should be, the intensity of the expected change is considered in the determination of this estimator by
\begin{align*}
    \tilde{\phi}_{k}^{(t)} = (1-p) \, \hat{\phi}_{k,v}^{(t-z_k^t):(t-1)} + p \, \hat{\phi}_{k,v}^{(t)}.
\end{align*}
Depending on the choice of this parameter, we are able to gradually alter the magnitude of change needed to be detected by the model. While a large $p$ only displays the most impactful changes, which are likely widely known, a smaller value for $p$ allows for experts on this topic to identify more niche changes.

\section{Evaluation}
\label{Evaluation}
\subsection{Data set}
To analyze the German political landscape, we use the protocols of plenary sessions of the German Bundestag. These were collected over the course of 72 years, starting from the first plenary session of the Federal Republic of Germany on the 7th of September 1949 until the the 3rd of June 2022 in the 20th legislative period. Each protocol can be downloaded from the website of the German Bundestag \cite{Bundestag} and is provided in an XML-format,  which contains, among other things, the date and entire plenary discussion in a text format. As one plenary session might contain multiple topics and points of discussion, we split these texts into smaller texts. Because there are a total of 4345 sessions we aim to split these texts automatically instead of manually and do so by splitting them into individual speeches using regular expressions. We also deleted the attachments and registers, as well as heckling and comments. This is an ongoing work but already provides better results than splitting the texts any arbitrary number of tokens or using the original plenary sessions as single documents. In total, the 4345 plenary sessions are split into 335\,065 documents. The distribution of documents by legislative period is displayed in the appendix in \autoref{tab:Dist}. The chunks of RollingLDA are adjusted to match the legislative periods, where each period is split into eight chunks (approximately two chunks per year).

\subsection{Study design}
For this study, we examined the different topic numbers $K=20, \ldots, 35$ each with $\alpha = \eta = 1/K$. For the RollingLDA we used the first legislative period as the initialization of the model. Starting from this, we modeled semi-annual minibatches, each using the last two years as memory. We applied the change detection algorithm with $p = 0.90, 0.91, \ldots, 0.95$ and $z_{\text{max}}=4$, i.e., for the detection of changes, a maximum of the previous 4 minibatches ($\sim$2 years $\approx$ \texttt{memory}) are taken as the reference period. If a change is detected for topic $k$ at time $t$, $z_k^{t+1}$ is set to 1, else to $\min\{z_k^t + 1, z_{\text{max}}\}$.

\subsection{Results}
\label{results}
Upon inspection of the results for the different parameters, we choose to present the findings for $K=30$ and $p=0.94$ in detail, yielding an interpretable number of detected changes while providing logical topics which can be analyzed separately from another and consistently over time. The following results serve as a proof of concept, as for a more fine-grained analysis in the future, a lower value of $p$ can be used. This way, the model will detect changes with a lesser impact on the topic, which will enable experts on German politics to identify changes that had an impact on German politics but may not be as well-known as the results we present here. All detected changes, corresponding top words, our interpretations and the results for other parameters can be accessed via the associated \href{https://github.com/K-RLange/zeitenwenden}{GitHub repository (K-RLange/Zeitenwenden)}.

\begin{figure*}[t]
\centering
\includegraphics[width=\textwidth]{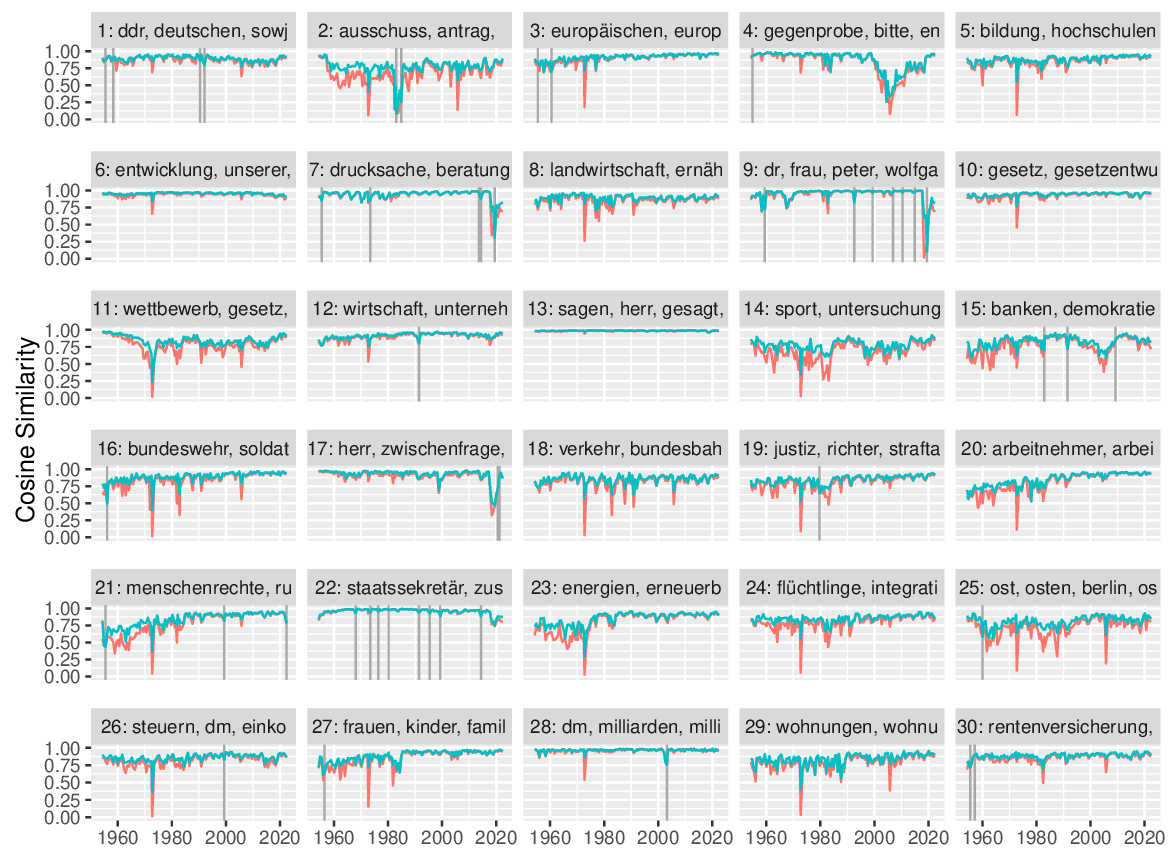} 
\caption{Observed similarity (blue), thresholds $q_k^t$ (red) and detected changes $C_k$ (vertical lines, gray) over the observation period for all topics $k \in \{1,\ldots,30\}$.}
\label{fig:res}
\end{figure*}

The changes are displayed in \autoref{fig:res}. The blue and red curves represent the observed similarities and the simulated quantile similarities, respectively. Each time the blue is below the red line, a change is detected as a gray vertical line.

The topics can be separated into political topics, which contain information about the current political discussions, and formality-topics which contain the names and titles of the parliament members as well as key words for common procedures, such as the voting process when deciding about a bill. While changes can be detected in either type of topics, changes in formality-topics will most likely not contain any information about the current political situation or discussion but rather about common political procedures or who is a current member of the parliament. Topic 9 for instance is a topic that is almost completely consisting of the names of parliament members. All 7 changes are detected at the start of a new legislative period, which is reasonable as new politicians join the parliament, but is not interesting for the sake of our analysis. Similarly, topics 4, 7 and 22 yield multiple changes that can be explained by a change in procedure or a different style of logging. Thus, we focus on the remaining 26 topics when looking for Zeitenwenden that were rooted in the topics of political discussions. We are able to link 22 of our 25 detected changes in relevant topics to interprable events. The remaining changes are caused by events that we were not able to interpret in retrospect.

\begin{figure}[H]
\centering
\includegraphics[width=0.48\textwidth]{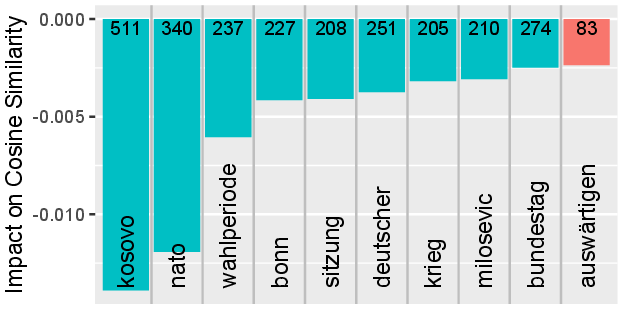}
\caption{Leave-one-out word impacts for topic 21 (1998-99), caused by the Kosovo war.}
\label{fig:kosovo}
\end{figure}

\begin{figure}[H]
\centering
\includegraphics[width=0.48\textwidth]{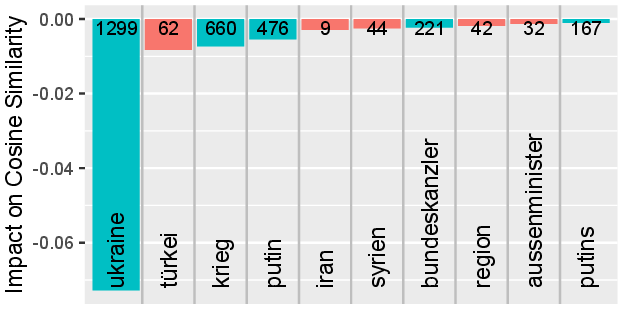}
\caption{Leave-one-out word impacts for topic 21 (2021-22), caused by the Russian-Ukrainian war.}
\label{fig:ukraine}
\end{figure}

Our model is able to detect some obvious events which affected political discourse such as the Russian-Ukrainian war (2022, topic 21, \citealp{Scholz}), the Covid 19 pandemic (2020-21, two changes in topic 17, \citealp{Pandemie}), the European financial crisis in 2008 (topic 15, \citealp{Hypo}), the introduction of the Euro as Germany's currency (2002, topic 28, \citealp{Euro}), the Kosovo-war (1999, topic 21, \citealp{Kosovo}), the German Reunification (1989-91, topics 1, 12, 15, \citealp{Reunification}), the founding of the Bundeswehr (1955, topic 16, \citealp{Bundeswehr}) and the Saar-referendum (1955, topic 3, \citealp{Saar}). Events like these had a long lasting impact on German society and politics and could be called \enquote{Zeitenwenden}. Interpreting the context of change is particularly easy, as the RollingLDA-model provides us with information about the overall topic of the change consistently over time. The Kosovo and Russian-Ukrainian war are for instance both detected in topic 21, which can be interpreted as the "war"{}-topic. To identify the exact reason for the change, we analyze the impact of each word using leave-one-out word impacts. Such word impact graphs are displayed in \autoref{fig:kosovo} and \autoref{fig:ukraine} for both mentioned wars. While most changes are caused by words being used more frequently due to a new event (blue bars), some are also caused by words that are used significantly less (red bars). The financial crisis of 2008 is a case in which the change is caused both by a change of focus within an event, as "ikb" was mentioned far less frequency, while words such as "krise" (crisis) started to emerge (see \autoref{fig:finanzkrise}).

While these were major changes which had a lasting impact on Germany, there are several smaller changes that are detected as well, such as the Bonn-Copenhagen declarations in 1955 recognizing the danish minorities in Schleswig-Holstein (topic 21, \citealp{Copenhagen}), the removal of the statute of limitations on murder (1979, topic 19, \citealp{Verjaehrung}) and the tax reform of 1998 (topic 26, \citealp{Steuer}).

\begin{figure}[h]
\centering
\includegraphics[width=0.23\textwidth]{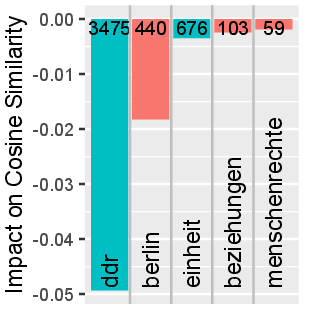}
\includegraphics[width=0.23\textwidth]{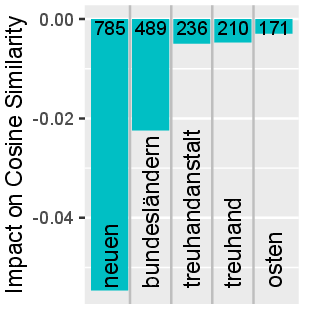}
\caption{Leave-one-out word impacts for topics 1 (1989-90) and 12 (1990-91) concerning East-Germany.}
\label{fig:ddr}
\end{figure}

Our results also enable us to reflect on the relationship between both German states BRD and DDR as well as BRD and western powers over the course of 40 years by interpreting the corresponding changes of major discussion points. West-Germany began major partnerships with western countries, such as the EGKS, a German-French cooperation that was founded according to the "Schuman-Plan" making the coal and steel-industries of both countries a European rather than a national matter (topic 3, \citealp{Schuman}). The Bonn-Paris conventions were also a result of both a closer connection towards western powers such as France and the troubled relationship between West-Germany and the eastern block (including East-Germany), as West-Germany became a member of NATO in 1955 (topic 1, \citealp{PariserVertraege}) and introduced a mandatory conscription in 1956 (topic 27, \citealp{Wehrpflicht}). All of this lead to the second Berlin-crisis in 1958, in which the Soviet Union demanded West-Berlin to become a free city rather than a part of West-Germany (topic 25, \citealp{Berlin}). Still, the NATO was not left unquestioned though, as the piece demonstrations in Bonn in 1981 against the NATO Double-Track Decision were a major discussion point in the Bundestag (topic 15, \citealp{Doppelbeschluss}). In 1990 West- and East-Germany unified. This is detected in several topics (1, 12, 15), as it was an long process which had a lasting impact in almost every political sectors, such as financial politics, inner politics, outer politics and many more. In 1991, a distinction was made between the \enquote{Neue Bundesländer} and \enquote{Alte Bundesländer}, denoting the parts of former East- and West-Germany after the unification. This was important as the parts of former East-Germany needed additional financial help to stabilize and reach the economical level of the western parts (topic 12). Ultimately, the usage of the word \enquote{DDR} decreases heavily in 1991 after both states had dissolved (topic 1, see \autoref{fig:ddr}).

\section{Summary}
\label{Summary}
To identify turning points in German political discourse, we analyzed plenary sessions of the German Bundestag from 1949 to 2021 using a change detection algorithm. This algorithm is based on a rolling version of the topic model LDA to create topics that are comparable across time. The changes detected reflect a significant change in the word distribution of the topics.

Our algorithm detects several meaningful changes over the course of the last 68 years of plenary discussions, such as key moments of the relationship between West- and East-Germany as well as world political events like the Russian-Ukrainian war, the Covid 19 pandemic and the financial crisis of 2008.

While these changes are identifiable as \enquote{true changes}, we do not know how many changes we missed, as major political discussions in the 21st century such as the refugee crisis in 2014 are not detected. This might be caused by a mixture-parameter that was chosen too restrictively or by the inability of the algorithm used to detect changes in topic distribution (see \autoref{fig:share}), as it is based on word distribution. Thus, topics that are suddenly a lot more relevant are not detected if the vocabulary used did not change. Identifying both would improve this analysis. Along with adjusting the mixture-parameter, this may enable a detailed analysis of Germany politics for experts on this topic. This can be further amplified by cleaning and using plenary sessions from 1867 to 1945, of East-Germany and of German state parliaments in addition to the Bundestag data set that we used here, as this would enable us to cover a broader spectrum of Germany's political discourse and history.

\section*{Acknowledgments}
The present study is part of a project of the Dortmund Center for data-based Media Analysis (\href{https://docma.tu-dortmund.de/}{DoCMA}) at TU Dortmund University. The work was supported by the Mercator Research Center Ruhr (MERCUR) with project number Pe-2019-0044. In addition, the authors gratefully acknowledge the computing time provided on the Linux HPC cluster at TU Dortmund University (LiDO3), partially funded in the course of the Large-Scale Equipment Initiative by the German Research Foundation (DFG) as project 271512359.

\bibliography{acl_latex}
\bibliographystyle{acl_natbib}

\appendix
\section{Additional Material} 

\begin{figure}[h]
\centering
\includegraphics[width=0.48\textwidth]{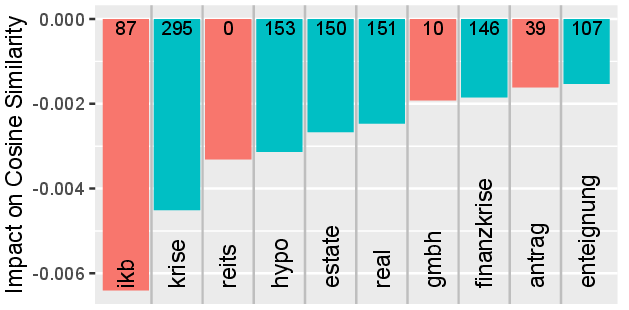}
\caption{Leave-one-out word impacts for topic 15 (2008-09), caused by the financial crisis.}
\label{fig:finanzkrise}
\end{figure}

\begin{figure*}[t]
\centering
\includegraphics[width=\textwidth]{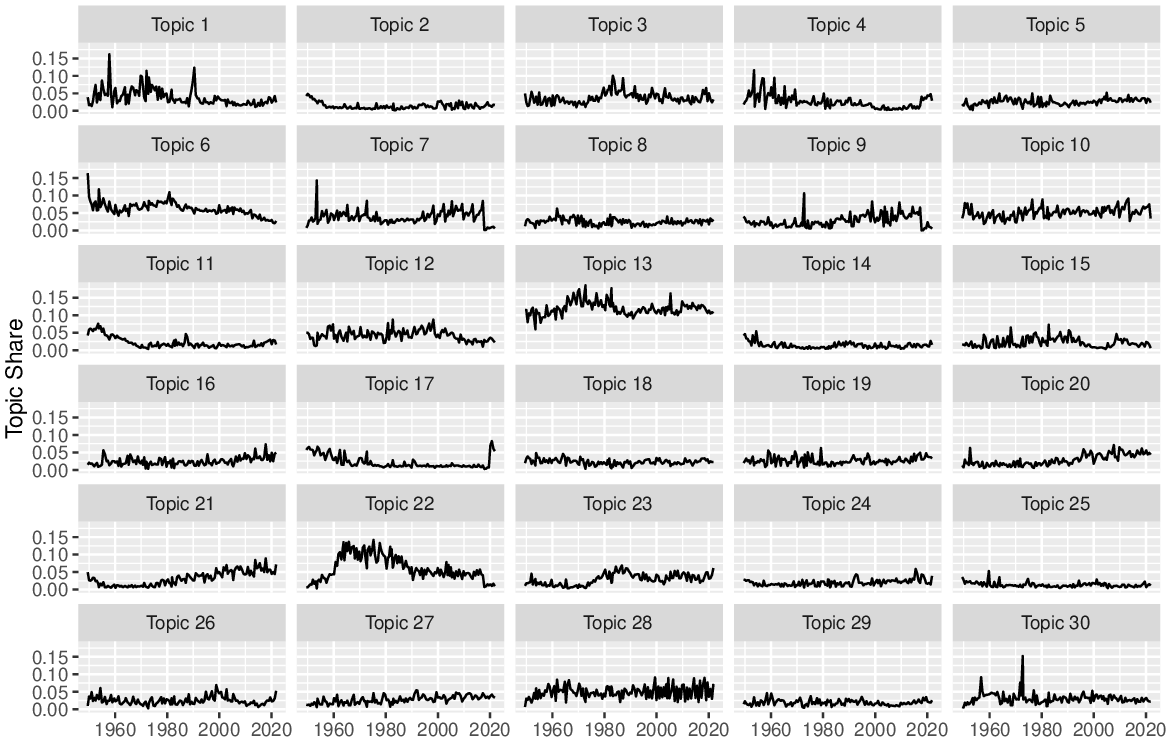} 
\caption{Topic shares per chunk: Relative number of assignments to the specific topic for a given time period.}
\label{fig:share}
\end{figure*}
\begin{table}[ht]
\centering
\caption{Approximate number of documents in relevant topics (see \autoref{results}) and detected changes for each legislative period.}
\label{tab:Dist}
\begin{tabular}{|cccc|}
  \hline
Period & Start date & Documents & Changes \\ 
  \hline
1 & 1949-09-07 & 16868 & NA \\ 
  2 & 1953-10-06 & 7963 & 7\\ 
  3 & 1957-10-15 & 7365 & 3\\ 
  4 & 1961-10-17 & 10900 & 0\\ 
  5 & 1965-10-19 & 17265 & 0\\ 
  6 & 1969-10-20 & 16540 & 0\\ 
  7 & 1972-12-13 & 19917 & 0\\ 
  8 & 1976-12-14 & 18199 & 1\\ 
  9 & 1980-11-04 & 10954 & 2\\ 
  10 & 1983-03-29 & 20812 & 1\\ 
  11 & 1987-02-18 & 20087 & 1\\ 
  12 & 1990-12-20 & 19780 & 3\\ 
  13 & 1994-11-10 & 20999 & 0\\ 
  14 & 1998-10-26 & 18582 & 2\\ 
  15 & 2002-10-17 & 13252 & 1\\ 
  16 & 2005-10-18 & 19867 & 1\\ 
  17 & 2009-10-27 & 25909 & 0\\ 
  18 & 2013-10-22 & 20752 & 0\\ 
  19 & 2017-10-24 & 25163 & 2\\ 
  20 & 2021-10-26 & 3891 & 1\\ 
   \hline
\end{tabular}
\end{table}

\end{document}